\documentclass[10pt,twocolumn,letterpaper]{article}
\usepackage{cvpr}
\usepackage{times}
\usepackage{graphicx}
\usepackage{amsmath}
\usepackage{amssymb}
\usepackage{epsfig}
\usepackage{epstopdf}
\usepackage[vlined,ruled,linesnumbered]{algorithm2e}
\usepackage{subcaption}
\cvprfinalcopy

\newcommand{\pname}[1]{{\textsc{MICIK}}{#1}}
\begin{document}

\title{MICIK: MIning Cross-Layer Inherent Similarity Knowledge for Deep Model Compression}

\author{Jie Zhang$^1$\thanks{Work was done during the first author's internship at Samsung Research America.}, Xiaolong Wang$^2$\thanks{Corresponding author.}, Dawei Li$^2$ , Shalini Ghosh$^2$,  Abhishek Kolagunda$^3$, Yalin Wang$^1$\\
$^1$Arizona State University, Tempe, AZ, USA \\
$^2$Samsung Research America, Mountain View, CA, USA\\
$^3$University of Delaware, Newark, DE, USA
}
\maketitle

\begin{abstract}
State-of-the-art deep model compression methods exploit the low-rank approximation and sparsity pruning to remove redundant parameters from a learned hidden layer. However, they process each hidden layer individually while neglecting the common components across layers, and thus are not able to fully exploit the potential redundancy space for compression. To solve the above problem and enable further compression of a model, removing the cross-layer redundancy and mining the layer-wise inheritance knowledge is necessary. In this paper, we introduce a holistic model compression framework, namely MIning Cross-layer Inherent similarity Knowledge (\pname{}), to fully excavate the potential redundancy space. The proposed \pname{} framework simultaneously, (1) learns the common and unique weight components across deep neural network layers to increase compression rate; (2) preserves the inherent similarity knowledge of nearby layers and distant layers to minimize the accuracy loss and (3) can be complementary to other existing compression techniques such as knowledge distillation. Extensive experiments on large-scale convolutional neural networks demonstrate that \pname{} is superior over state-of-the-art model compression approaches with 16X parameter reduction on VGG-16 and 6X on GoogLeNet, all without accuracy loss.
\end{abstract}


\section{Introduction}
Deep learning is the primary driving force for recent breakthroughs in various computer vision applications, such as object classification~\cite{krizhevsky2012imagenet}, semantic segmentation~\cite{turaga2010convolutional}, image captioning~\cite{im2txt2}, etc. However, deploying deep learning models on resource constrained mobile devices is challenging due to high memory and computation cost although most mobile vision applications can benefit from the advantages of on-device deployment such as low latency, better privacy and offline operation~\cite{deeprebirth,howard2017mobilenets}. Motivated by this demand, researchers have proposed deep model compression methods to remove redundancy in the learned deep models, i.e., reducing both computing and storage cost with minimum impact on model accuracy.

\begin{figure}[t]
\centering
\vspace{-1em}
\includegraphics[height=6cm]{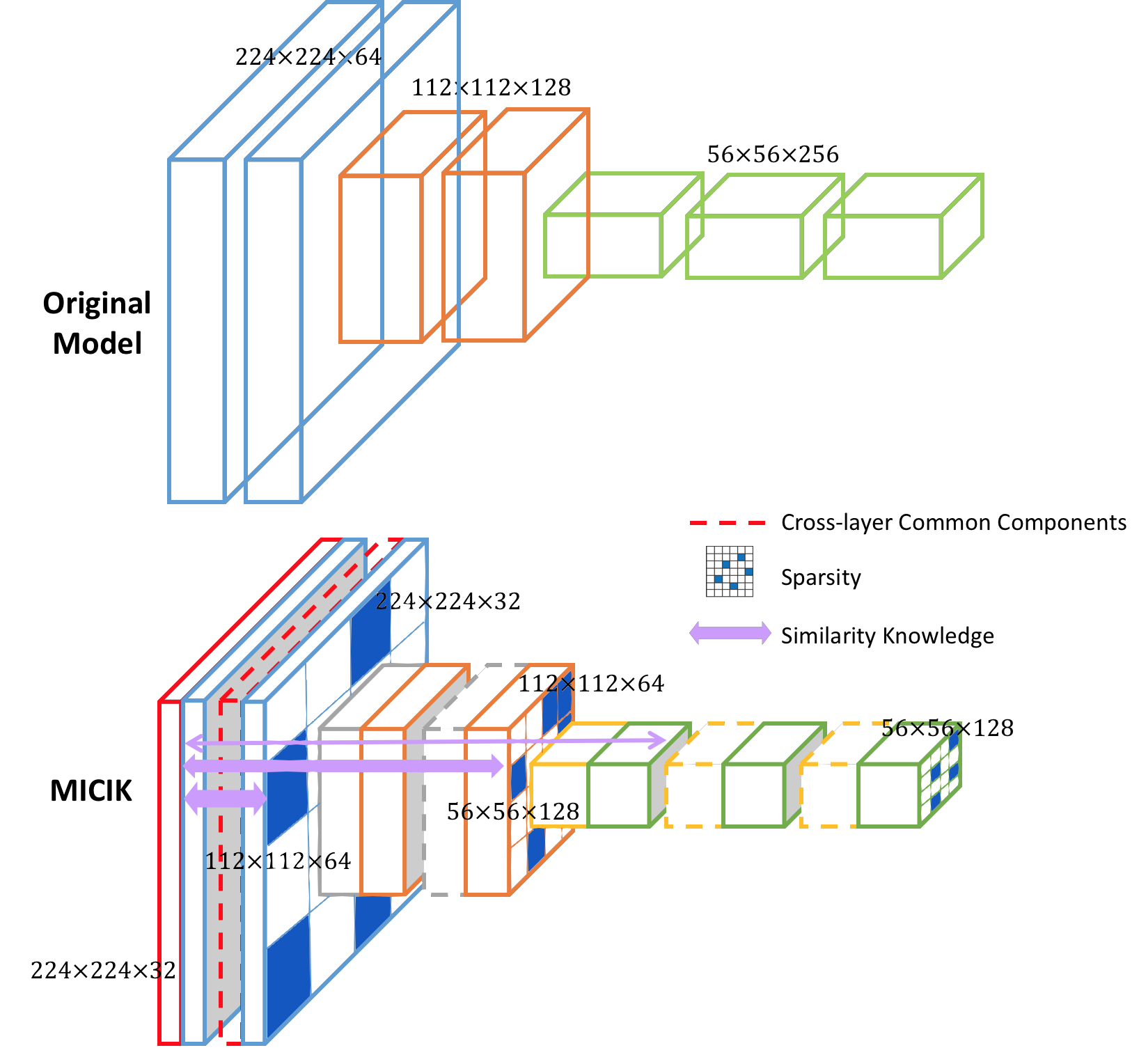}
\caption{An illustration of the model compression architecture of \pname{}. \pname{} obtains the low-rankness and sparsity of the original network structure. Several layers share the common components (dashed line) for higher compression rate and solid lines (of a different color) represent the exclusive weight components of each layer. Besides, the nearby layers (different colors for different layers) have higher inherent similarity than distant layers (thicker arrow shows higher similarity). } 
\label{fig:1}
\vspace{-1em}
\end{figure}

One of the most widely used model compression approaches is pruning. The general idea of pruning is to directly remove redundant parameters based on certain pre-defined strategies. In an early work, Han \etal~\cite{han2015learning} directly applied a hard threshold on the weights to cutoff network's unimportant connections. He \etal~\cite{he2017channel} pruned the network by pruning each layer with a LASSO regression based channel selection and least square reconstruction. 
Another prevalent approach for model compression is low-rank approximation which decomposes a large weight matrix into several small matrices and has been successfully applied to both convolutional and fully-connected layers \cite{tai2015convolutional,kim2015compression}. Low-rank approximation generally provides better initialization for fine-tuning than sparsity pruning and thus has the advantage of greatly reduced retraining time. However, low-rank decomposition cannot fully represent the knowledge in the original layers since some important information is distributed outside the low-rank subspace. 
Recently, as observed by Yu \etal~\cite{yucompressing}, combining low-rank approximation with sparse matrix decomposition gives better compression rates than using only either method individually.

In this paper, we argue that existing model compression approaches only apply within each layer separately and neglect the correlation of different layers. However, a deep learning model is usually composed of multiple layers where the parameters are learned in a sequential structure. Therefore there must be some common components shared among weight matrices of different layers, especially between adjacent layers (e.g., the first layers extract low-level features such as edges). Mining these shared common representations enables further compression of the deep model. Motivated by the previous work~\cite{yucompressing} and the information sharing assumption, we propose a holistic model compression method, namely \pname{}, to decompose layers' weight matrices using low-rank and sparsity component with integration of cross-layer mining scheme. 
Specifically, \pname{} considers mining the shared information across layers while preserving inherent similarity and formulates it as a multi-task learning (MLT)~\cite{evgeniou2004regularized} problem where both intra-layer and inter-layer feature representations are embodied together. In addition, learning the correlation among layers while treating all layers equally can distort the original deep sequential structure, especially for the layers with a large gap. Therefore, we incorporate an inherent knowledge mining scheme into the proposed pipeline to maximize the similarity of nearby layers and the dissimilarity of distant layers during the compression. 

Our main contributions can be summarized as follows. 
First, unlike previous layer-wise model compression approaches which perform optimization based only on intra-layer correlations, we propose a holistic deep model compression framework that mines both common and individual weight components' correlation simultaneously. To the best of our knowledge, this is the first compression work that considers using shared common weight components across multiply layers. This significantly increases the compression capacity in the deep learning model matrices.
Second, to maintain the consistency of deep feature learning, our framework also employs inherent similarity knowledge to better learn the network structure, which will make the correlation of adjacent layers closer and the correlation of distant layers farther. By adding this term, the compression rate may be further improved by about $10\%$ without accuracy loss of the compressed model. Furthermore, an efficient optimization method is introduced. Third, the proposed approach is also complementary to other existing compression techniques such as knowledge distillation~\cite{hinton2015distilling} (KD) which can further improve the model compression performance. We have evaluated the proposed approach on two popular CNN architectures: GoogLeNet and VGG-16. Extensive evaluation demonstrate that the superiority of the proposed model compression algorithm over state-of-the-arts and significantly improves the compression rate.

\section{Related Work}
Recently, deep model compression has drawn great interests from AI researchers. We summarize previous compression strategies into four different categories.
\subsection{Sparsity based model compression}
Sparsifying parameters in learned neural network layers is an intuitive approach for compression. One general way for sparsification is pruning, which iteratively removes network parameters that have the least impact to the final prediction accuracy. For example, Han et al.~\cite{han2015learning} proposed using a hard threshold formed by the product of a scalar and the standard deviation of the weights. Following this work, filter-level pruning methods~\cite{luo2017thinet} have been introduced which enables a more structured pruning strategy. Most recent, He \etal~\cite{he2018amc} proposed using AutoML for model compression, which gave larger acceleration rate than others. The problem with pruning based compression is that it cannot guarantee a good initialization for retraining and thus requires tedious iterative processing.

\subsection{Low rankness based model compression}
Matrix quantization trades network representation precision of weights and activations for computational throughput. Xue \etal~\cite{xue2013restructuring} used low-rank approximation to reduce parameter dimensions which saves storage and improves training and testing times. Denton \etal~\cite{denton2014exploiting} explored matrix factorization methods for speeding up CNN inference time. These works showed it can be exploited to speed up CNN testing time by as much as 200\% while keeping the accuracy drop within 1\% of the original model. To avoid the accuracy loss and compensate the lost information for low-rankness, Yu \etal~\cite{yucompressing} integrated both sparse and low-rank decomposition so that both smooth and spiky components can be preserved. All those methods are limited to compressing single-layer individually, while our method exploits the cross-layer relation for further compression. Zhang \etal~\cite{zhang2018dynamically} used low-rank approximation to compress Recurrent Neural Network. 

\begin{figure*}[t]
\centering
\vspace{-1em}
\begin{subfigure}{.51\textwidth}
  \centering
  \includegraphics[height=3cm]{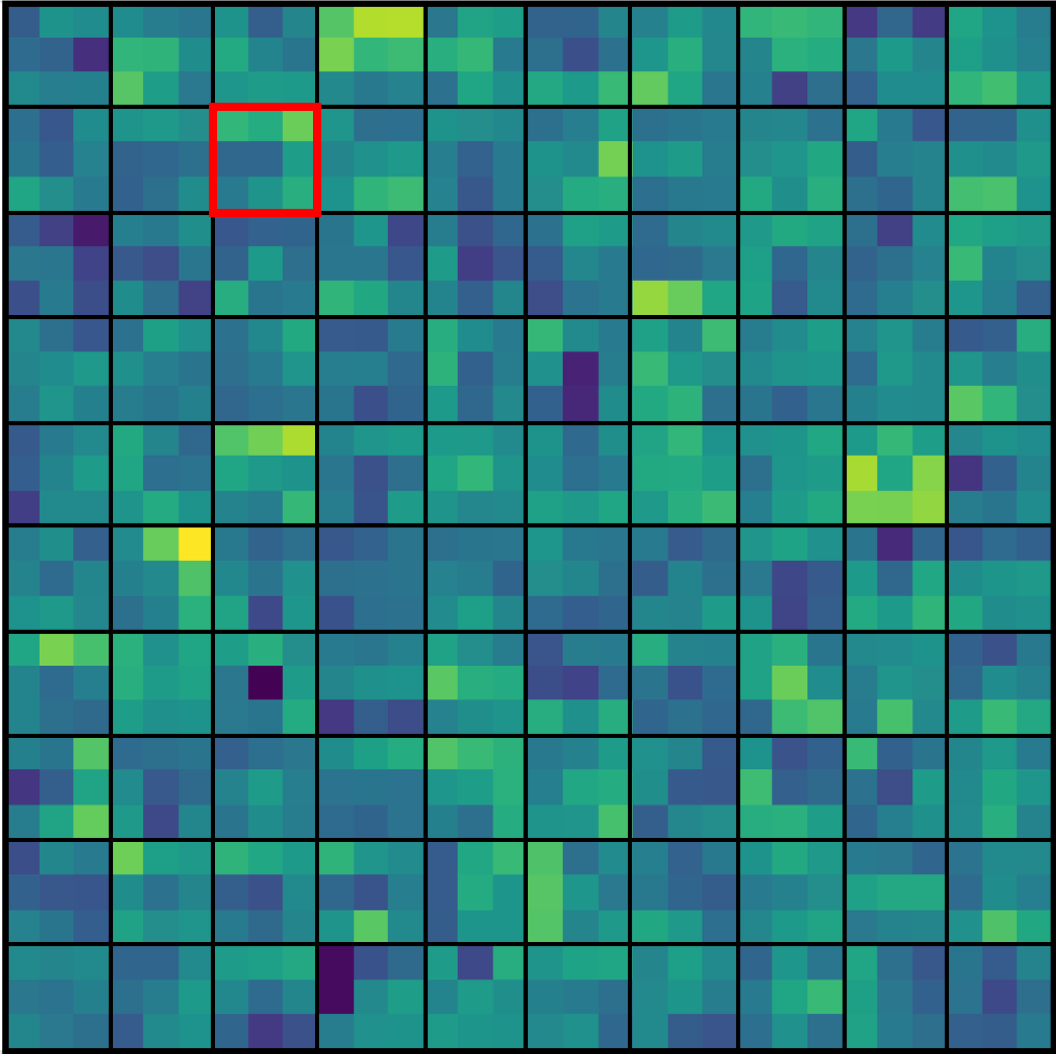}
  \includegraphics[height=3cm]{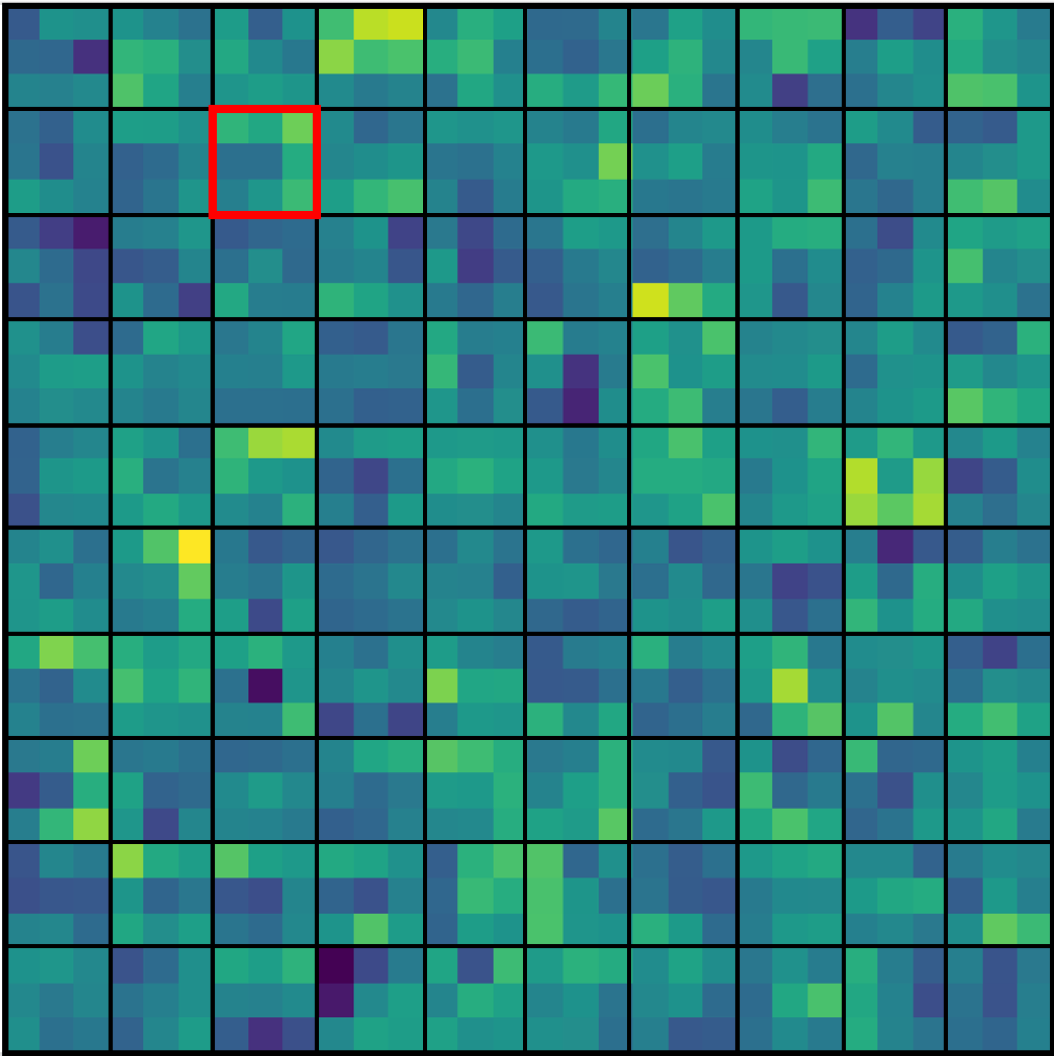}
  \caption{Layers \emph{inception\_3a} and \emph{inception\_3b}}
\end{subfigure}%
\begin{subfigure}{.5\textwidth}
  \centering
  \includegraphics[height=3cm]{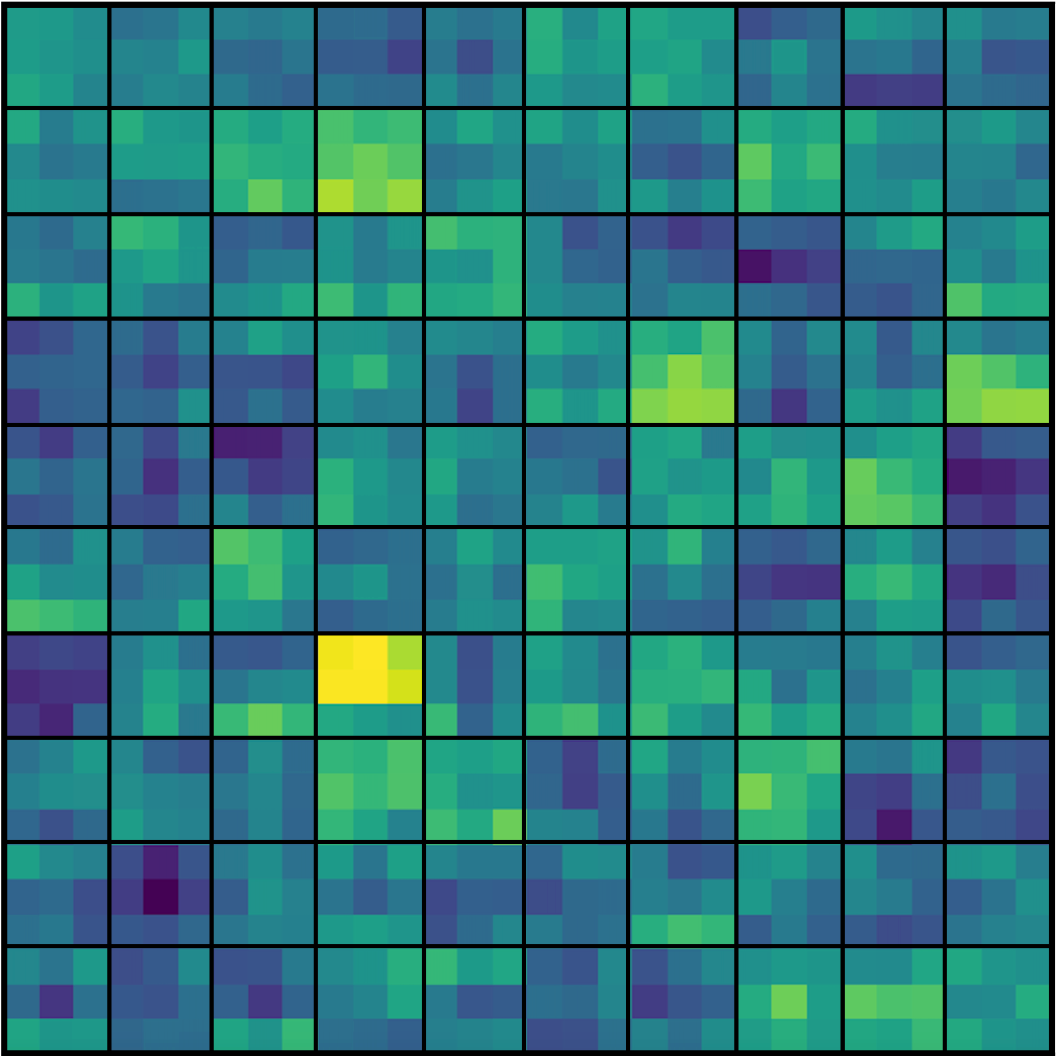}
  \includegraphics[height=3cm]{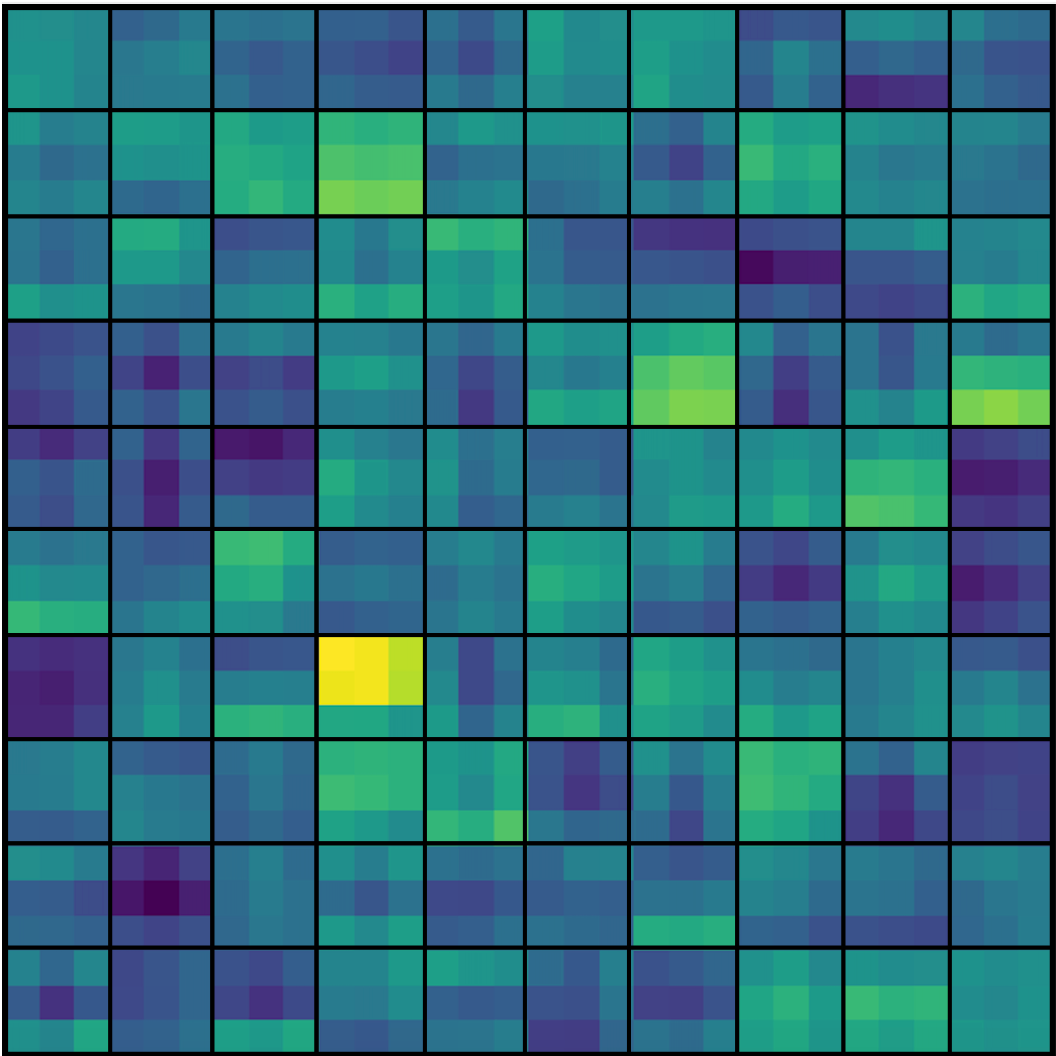}
  \caption{Layers \emph{inception\_3a} and \emph{inception\_5b}}
\end{subfigure}
\caption{The closest 100 corresponding filters ($3\times 3$) from close
two layers (a) and distant two layers (b) of GoogLeNet.}
\label{fig:2}
\vspace{-1em}
\end{figure*}

\subsection{Knowledge transfer methods}
knowledge distillation (KD) transfers knowledge from one or several large pre-trained teacher networks to a small student network which can achieve comparable capability of the teacher networks during inference. Hinton \etal~\cite{hinton2015distilling} trained a small student network to match the soft targets of a cumbersome teacher network by setting a proper temperature parameter. Chen \etal~\cite{chen2017darkrank} introduced cross sample similarities and brought the learning to rank technique to model compression and acceleration. Our approach is compatible with KD to achieve higher compression rate.

\subsection{Quantization based model compression}
    Quantization methods group weights with similar values or reduce the number of required bits to reduce the number of parameters. Gong \etal~\cite{gong2014compressing} applied vector quantization to the parameter values. Han \etal~\cite{han2015deep} pruned the unimportant connections and retrained the sparsely connected networks, then quantized the link weights using weight sharing and Huffman coding to further reduce the model size. Binary-weight neural networks such as BinaryNet~\cite{courbariaux2016binarized} and XNor-Net~\cite{rastegari2016xnor} use one binary bit to represent each weight parameter while maintaining the model accuracy. Our approach operates in a more principled way for compressing neural networks and this guarantees a faster converge speed during the re-training stage.

\section{Method}
In this section, we will introduce the proposed \pname{} framework in detail. We first review the basic low-rank and sparse decomposition for a single layer. Next, we illustrate the motivation of cross layer compression and present the algorithm which simultaneously mines both the individual and common weight components. Specifically, \pname{} takes into consideration the inherent similarity knowledge across successive layers to formulate the objective function. Finally, we introduce the optimization scheme.

\subsection{Single-layer Compression}
For the single-layer compression, the output of a convolutional or a fully connected layer can be obtained by 
\begin{equation}
y = Wx,
\label{eq:in}
\end{equation}
where $x \in \mathbb{R}^p$ is the input feature vector, $W \in \mathbb{R}^{n \times p}$ is the weight matrix of a convolutional or fully connected layer and $y \in \mathbb{R}^n$ is the output response. Assume $L$ is the approximation of $W$ in low rank subspace.   
$L$ can be represented as the product of two smaller matrices as $L = UV \in \mathbb{R}^{n\times p}$, where $U \in \mathbb{R}^{n\times m}$ and $V \in \mathbb{R}^{m\times p}$ ($m < n$ and $m < p$). Therefore, we have the following model:
\begin{equation}
\label{eq:1}
\begin{split}
\min_L \quad \frac{1}{2}||W - L||_F^2,\\
s.t. \quad rank(L) \leq m,
\end{split}
\end{equation}
where $rank(L)$ represents the rank of $L$.
However, if the weight matrix is represented only in the low-rank subspace, some important sparse information could be lost.

To recover the information loss, Yu et al. \cite{yucompressing} proposed to incorporate an additional sparse matrix $S \in \mathbb{R}^{n\times p}$. Thus, to compress the weight matrix $W$, Eq.~(\ref{eq:1}) is reformulated as follows:
\begin{equation} 
\label{eq:2}
\begin{split}
\min_{L, S} \quad &\frac{1}{2}||W- L - S||_F^2,\\
s.t. \quad &rank(L) \leq m, card(S) \leq q,
\end{split}
\end{equation}
where $card(S)$ denotes the cardinality of matrix $S$.
The compression rate achieved using this method is $\frac{m(n+p) + q}{np}$.

If we sequentially apply Eq~\ref{eq:2} to several layers prior to a retraining, the approximation error of each layer will be accumulated and it will be hard for the compressed model to converge. Therefore, we add an asymmetric data reconstruction term in Eq.~\ref{eq:single-layer} to reduce the accumulated error from Eq.~\ref{eq:2}. This asymmetric term ensures the optimal approximation of weight matrices in a given layer and avoids the abrupt accuracy loss during the compressing process which can help speed up the convergence.
\begin{equation}
\begin{split}
\min_{L, S} &\frac{1}{2}||Y - (L + S)X||_F^2,\\
s.t. & \frac{1}{2}||W - L - S||_F^2 \leq \epsilon, rank(L) \leq m, card(S) \leq q,
\end{split}
\label{eq:single-layer}
\end{equation}

\subsection{Mining Cross-layer Model Compression}
However, Eq.~(\ref{eq:single-layer}) does not consider the relation among different layers which may share common components for further model compression. We observe that different layers of a deep neural network have shared common components (similarity). This becomes our initial motivation for exploring cross-layer model compression. Fig.~\ref{fig:2} shows corresponding filters in different $3\times 3$ convolutional layers of GoogLeNet. Red boxes in Fig.~\ref{fig:2} (a) show the common components of nearby layers. We say two filters, each from a different layer, are correspondences if they are each other’s nearest filter, measured by $L_2$ similarity. It can be seen that for both close layers (Fig.~\ref{fig:2} (a)) and distant layers (Fig.~\ref{fig:2} (b)), the correspondences formed between layers appear very similar. It is worth noting that our algorithm does not rely on the correspondence definition and we demonstrate that different neural architectures can be well handled by the proposed solution in the experiments. 

To fully exploit such potential redundancy, the idea of Multi-Task Learning (MTL)~\cite{evgeniou2004regularized} can be incorporated. MTL was proposed to improve learning performance by learning multiple related tasks simultaneously and training tasks in parallel by using a shared representation. Based on MTL, we redefine Eq.~(\ref{eq:single-layer}) and propose an algorithm to compress weight matrices from different layers simultaneously. 

Given the weight matrices from $T$ different layers $\{W_1,W_2, \cdots ,W_T\}$, the proposed cross-layer compression can be formulated as follows: 
\begin{equation}
\label{eq:3}
\begin{split}
\min_{U, V, S}\ &\frac{1}{2}\sum_{t=1}^T||Y_t - (L_t + S_t)X_t||_F^2 + \lambda\sum_{t=1}^T||W_t - L_t \\- & S_t||_F^2,\ \ s.t.\ \ rank(L_t) \leq m_t, \  card(S_t) \leq q,
\end{split}
\end{equation}
where $\lambda$ is an non-negative parameter and provides a trade-off between the loss function and the asymmetric term. In Eq.~(\ref{eq:3}), MTL is used directly to learn $L_t = U_tV_t$ individually in parallel, but does not make clear how the redundancy among layers can be exploited. Therefore, to further increase the compression rate, \pname{} is proposed to use common and individual components to compress across multi-layer simultaneously. 

For the weight matrix $W_t$ of a particular layer, our goal is to learn the weight components $V_t$ which are composed of two parts: $V_t = [\hat{V_t}, \bar{V_t}]$ where $\hat{V_t}\in \mathbb{R}^{\hat{m} \times p} $, $\bar{V_t}\in\mathbb{R}^ {\bar{m}_t \times p}$ and $\hat{m}+\bar{m}_t=m_t$. $\hat{V_t}$ is the common component among different layers and $\hat{V_1}= \cdots=\hat{V_T}$ while $\bar{V_t}$ is different from each other and only learned from the corresponding matrix $W_t$. The objective function can be reformulated as follows:
\begin{equation}
\label{eq:multi-layer}
\begin{split}
&\min_{U, V, S}\frac{1}{2}\sum_{t=1}^T||Y_t - (U_t[ \hat{V_t}, \bar{V_t}] + S_t)X_t||_F^2 + \lambda\sum_{t=1}^T||W_t - L_t \\&- S_t||_F^2,
s.t. \hat{V_1} =\cdots=\hat{V_T}, rank(L_t) \leq m_t,card(S_t) \leq q.
\end{split}
\end{equation}
 
In Eq.~(\ref{eq:multi-layer}), we can compress several convolutional layers' weight matrices together. We used a common components matrix $\bar{V}$ across multiple layers to further compress the model and the total parameters for each layer will become $(\bar{m_t} + \hat{m}/T)(n+p) + q$ instead of $m_t(n+p)+q$. 

\subsection{Mining Inherent Similarity Knowledge}
The layers in a deep neural network have an ordinal relationship. Layers close to the input extract low-level image features (e.g., edges) while top layers close to the output extract high-level features with semantic meanings, thus there are different degrees of similarity between nearby layers and distant layers. 

\begin{figure*}[t]
\centering
\begin{subfigure}{.5\textwidth}
  \centering
  \label{fig3:a}
  \includegraphics[height=4cm]{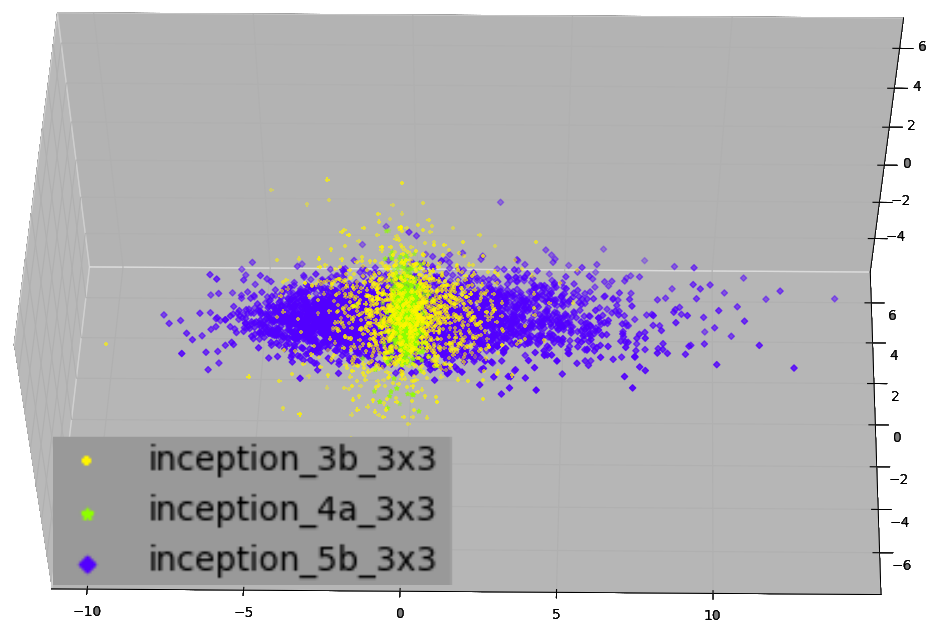}
  \caption{Layers \emph{inception\_3b, 4a} \\and \emph{inception\_5b}}
\end{subfigure}%
\begin{subfigure}{.5\textwidth}
\label{fig3:b}
  \centering
  \includegraphics[height=4cm]{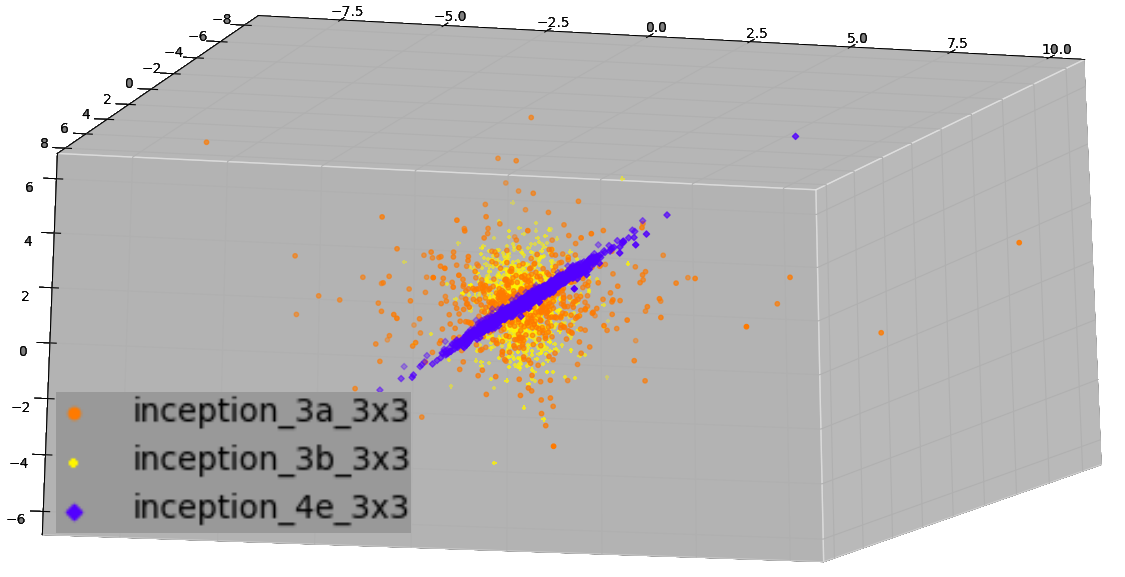}
  \caption{Layers \emph{inception\_3a, 3b} \\and \emph{inception\_4e}}
\end{subfigure}
\caption{The similarity distribution of kernels visualized using the first three dominant principal components of GoogLeNet.}
\label{fig:PCA}
\vspace{-1em}
\end{figure*} 

Fig.~\ref{fig:PCA} shows distributions two examples of groups of three layers from GoogLeNet. These three layers are chosen such that two of them are consecutive layers with the same filter size and the third one is the farthest layer with filters of the same size ($3\times 3$). We use principal component analysis (PCA)~\cite{wold1987principal} to project them onto a 3D-subspace for visualizing the distributions. The three selected layers are sorted by their depth in the network and we perform PCA to project the filters onto the three most dominant principal components A for the first layer. For the second and third layer, we apply PCA on the filter kernels and find the principal components B that are most similar (cosine distance) to A and plot them in the 3D subspace. We use different colors to represent each layer. We can see that the distributions of inception 3b, 4a (Fig.~\ref{fig:PCA} (a)) share more similarities than inception 5b and inception 3a, 3b (Fig.~\ref{fig:PCA} (b)) share more similarities than inception 4e. We observed that nearby layers are more likely to have similar distributions when compared to layers that far apart in depth and this motivates us to add constraints between codes learned for successive layers. And this means the nearby layers would have more similarity on their individual features than layers that are far away from each other. Therefore, we used a weighted function to emphasize the inherent similarity knowledge between two different layers:
$\theta_{ij}=\left\{
\begin{array}{c}
\frac{1}{j - i} \quad i < j;\\
0 \quad i \geq j.
\end{array}
\right.$

The function $\theta_{i,j}$ is used to penalize the distance between two layers so that it emphasizes the inherent similarity of two layers, i.e., the nearby layers from learning layer-specific components $\bar{V_t}$ with high similarity and distant layers with high disparities. For example, the distant layers $(W_2, W_4) (\theta_{2,4} = 1/2)$ have larger differences than nearby layers $(W_2, W_3) (\theta_{2,3} = 1)$, we use a smaller $\theta_{2,4}$ to yield the smaller similarity of two distant layers so that the distant layers $\bar{V_2}$ and $\bar{V_4}$ will yield larger differences than nearby layers.  

The final objective function of the \pname{} compression algorithm can be formalized as follows:
\begin{equation}
\label{eq:5}
\begin{split}
\min_{U, V, S} &\sum_{t=1}^{T} \frac{1}{2}||Y_t - (U_t[\hat{V_t},\bar{V_t}]+S_t)X_t||^2_F+ \\ &\lambda\sum_{t=1}^T||W_t - L_t - S_t||_F^2 
-\lambda_{\theta}\sum_{j=1}^T\sum_{i=1}^T\theta_{ij}||\bar{V_i} - \bar{V_j}||_{2}^2,\\
s.t.& \ \hat{V_1} =\cdots=\hat{V_T}, \  rank(L_t) \leq m_t, \ card(S_t) \leq q.
\end{split}
\end{equation}

To optimize Eq~\ref{eq:5}, we alternately optimize $U, V$ and $S$. When updating $U, V$, to avoid excessive matrix inversions, we use ~\cite{zhou2013greedy} which reduces the time complexity from $O(\min(n^2p, np^2))$ to $O(2npm)$ ($m < \min(n, p)$). When updating $S$, to avoid calculating the inverse of Hessian matrix $\mathbf{H}$, we use the diagonal element of $\mathbf{H}$ because $\mathbf{H}$ would be close to a diagonal matrix when the columns of U have low correlation. Thus, we reduce the time complexity from $O(m^2)$ to $O(m)$ for updating $S$. 

\subsection{Optimization Analysis}
\label{sec:1}
In this section, we explain the update procedure for the low-rank structure $L_t$ and sparse structure $S_t$ to minimize Eq.~(\ref{eq:5}). The whole process of updating $U$, $V$ and $S$ is summarized into Algorithm~\ref{alg:1}. We use an alternating updating method 
to update $U_t$, $V_t$ and $S_t$ and run it for $k$ iterations~(Alg.~\ref{alg:1}). More specifically, the objective function in Eq.~\eqref{eq:5} is a non-convex problem due to the non-convexity of the components $card(S)$ and $rank(L)$. We use the convex relaxation technique~\cite{boyd2004convex} to solve the above problem. 

First, we update $U$ by fixing $V$ and $S$, we use the same updating rule (QR decomposition) as~\cite{yucompressing} for each input layer $t$:
\begin{equation}
U_t^k = Q, QR(B^k(V_t^k)^T) = QR; V_t^k = Q^T(B^kA^{\S}),
\label{eq:QR}
\end{equation}
where $A = \lambda I + \frac{1}{n}X_tX_t^T$ and $B^k = \lambda(W_t - S_t^{k-1}) + \frac{1}{n}(Y_tX_t^T - S_t^{k-1}X_TX_t^T).$ $A^\S$ is the Moore-Penrose Pseudoinverse~\cite{albert1972regression} of $A$.

\begin{algorithm}[t]
\caption{\pname{}}
\label{alg:1}
\SetKwData{Left}{left}\SetKwData{This}{this}\SetKwData{Up}{up}
\SetKwFunction{Union}{Union}\SetKwFunction{FindCompress}{FindCompress}
\SetKwInOut{Input}{Input}\SetKwInOut{Output}{Output}
\Input{$W$ from DNN, $X, Y$, epoch $\kappa$, layer $T$, $\lambda$, $m$, $q$ and initial $\hat{V}=V_0$}
\Output{ $U, V$ and $S$}
\Begin{
   \For{ $k=1 \rightarrow \kappa$ }{
   \For{ $t=1 \rightarrow T$ }{
     For each input weight matrix $W_t$;\\
     Update $U_t^k$ by QR decomposition~\cite{yucompressing} as Eq.~\eqref{eq:QR};\\
     Update $\hat{V}^k_t$: $\hat{V}^k_t=V_0$;\\
     Update the $\hat{V_t^k}$ and $\bar{V_t^k}$ by Eq.~\eqref{eq:updateV};\\
   Calculate $\theta$ function and add $\lambda_{\theta}\sum_{i=1}^{T}\sum_{j=1}^T\theta_{ij}||\bar{V_i}-\bar{V_j}||$ term;\\
   Update $S_t^{k}$ by Eq.~\eqref{eq:updateS}; \\
   $V_0=\hat{V}^{k+1}_t$;\\
   }
  }
}
\end{algorithm}

Then, we update $V$ by fixing $U$ and $S$ and we need to consider two parts of $V$: $\hat{V}$ and $\bar{V}$. Let $f: \mathbb{C} \rightarrow \mathbb{R}$, where $\mathbb{C} \subseteq \mathbb{R}^{n\times p}.$ The convex envelope of $f$ on $\mathbb{C}$ is defined as the largest convex function $g$ such that $g(x) < f(x), \forall x \in \mathbb{C}$. The trace norm (nuclear norm) has been known as the convex envelop of the function of rank~\cite{fazel2001rank}: 
\begin{equation}
||L||_\ast \leq rank(L), \forall L \in \mathbb{C} = \{L | ||L||_2 \leq 1\}.
\end{equation}
Therefore, the equivalent objective function of Eq~\eqref{eq:5} on updating $V$ will become:
\begin{equation}
\begin{split}
&\min_{\hat{V}_t, \bar{V}_t} \frac{1}{2}  \sum_{t=1}^T||Y_t-(U_t[\hat{V}_t, \bar{V}_t] + S_t)X_t||^2_F+\lambda\sum_{i=1}^T||W_t - \\& L_t-  S_t||_F^2 - \lambda_{\theta} \sum_{i=1}^T\sum_{j=1}^T\theta_{ij}||\bar{V_i} - \bar{V_j}||_{2}^2, s.t. ||L_t||_\ast \leq m_t.
\label{eq:6}
\end{split}
\end{equation}
In Alg.~\ref{alg:1} (line 5), to construct $L$ with rank $m$, instead of updating $m$ columns/rows at all iterations, we use a greedy selection~\cite{zhou2013greedy} to update the $U$ and $V$. Specifically, we initiate a small rank $(m = 1)$ to start, then select extra $\triangle m$ columns/rows and concatenate them into $V$ at each iteration. Such greedy method is a warm-start for the higher rank optimization and ensures the faster computation compared to updating m columns/rows at all iterations. Thus, we still need to constrain the rank of $L$ in Eq.~\ref{eq:6}. 

We use $V_t^k$ obtained from Eq.~\eqref{eq:QR} as the initialized $V_t^0$ and we perform one step stochastic gradient descent (SGD)~\cite{bottou2010large} to update the weight components $\hat{V}^{k+1}_t$ and $\bar{V}^{k+1}_t$. The procedure of optimization takes the form of
\begin{equation}
[\hat{V}^{k}_t,\bar{V}^{k}_t] = [\hat{V}^{k-1}_t,\bar{V}^{k-1}_t] - \frac{R}{H_t^{k}}U_t^{k} - S_t^{k},
\label{eq:updateV}
\end{equation}
where $H$ is the Hessian matrix of $V_t$. $R$ stores $U_tV_t +S_t - W_t$ of layer $t$ and $R = \Omega([\hat{V}^{k-1}_t, \bar{V}^{k-1}_t], U_t^{k-1}, S_t^{k})-W_t.$ Here, $\Omega(A, B, I)$ is a matrix multiplication function and has three input parameters ($A$, $B$ and $I$ denotes matrices) and $\Omega(\cdot) = AB + I$. 

After we update $V_t^{k}$ using Eq.~\eqref{eq:updateV}, we can now calculate $\theta_{ij}$, and then easily add the result of the third term of Eq.~\eqref{eq:6} on $\bar{V_t^{k}}$ at the end of $k$th updating $V$ iteration.

Last, we update $S_t$ by fixing $U_t$ and $V_t$. $l_0$-norm (cardinality) (the number of non-zero entries) is normally used to control the sparsity of the matrix, and the optimization problem of updating $S$ is equivalent to the follows:
\begin{equation}
\begin{aligned}
\min_{S} \quad &\frac{1}{2} \sum_{t=1}^{T} ||Y_t - (U_t[\hat{V_t},\bar{V_t}]+S_t)X_t||^2_F \\+& \lambda\sum_{t=1}^T||W_t - L_t - S_t||_F^2 + \lambda_2\sum_{t=1}^T||S_t||_0,
\end{aligned}
\end{equation}
where $\lambda_2$ is an non-negative parameter to control the sparse regularization. However, solving $l_0$ norm is NP-hard~\cite{natarajan1995sparse}, and $l_0$ regularization is generally intractable. Therefore we use convex relaxation techniques~\cite{boyd2004convex}: $l_1$ norm is known as the convex envelope of the $l_0$-norm, and thus $l_1$-norm is used instead of $l_0$ norm:
\begin{equation}
\begin{split}
\label{eq:8}
\min_{S} \quad &\frac{1}{2} \sum_{t=1}^{T} ||Y_t - (U_t[\hat{V_t},\bar{V_t}]+S_t)X_t||^2_F \\+& \lambda\sum_{t=1}^T||W_t - L_t - S_t||_F^2 + \lambda_2\sum_{t=1}^T||S_t||_1.
\end{split}
\end{equation}

Given~\ref{eq:8}, the corresponding $l_1$ norm regularized convex optimization problem can be efficiently solved. 
We calculate the gradient $h$ based on Eq.~(\ref{eq:updateS}), and then update the model $S_t^{k}$ based on $h$. The calculation of $h$ and $S_t^{k}$ follows the equations:
\begin{equation}
\begin{split}
\label{eq:updateS}
&h = (U_t[\hat{V}^{k-1}_t,\bar{V}^{k-1}_t]_j + S_t^{k-1})^T (\Omega( [ \hat{V}^{k-1}_t, \bar{V}^{k-1}_t], U_t^{k-1},\\ &S_t^{k-1}) -W_t),
\quad S_t^{k}=\Gamma_{\lambda_2 }( S_t^{k-1}-h),
\end{split}
\end{equation}
$\Gamma$ is the soft thresholding shrinkage function \cite{combettes2005signal}. 


\begin{table}[t]
\centering
\vspace{-1em}
\caption{Compression statistics of \pname{} on GoogLeNet. $\#W(O)$: total parameters of original model. $\#W(C)$: total parameters of compressed model. R: compression rate.}
 \begin{tabular}{c|c|c|c|c}
\hline
Layers & $\#$ W (O) & $\#$W (C)& R &GreBdec\\ \hline
conv1\_1 &9k&4.5k&2X&1X \\
conv2 &115k&57K&2X&1.2X\\
inception\_3a &164k&33k&5X&4.8X\\
inception\_3b &389k&77k&5X&4.5X\\
inception\_4a &376k&75k&5X&4.5X\\
inception\_4b &449k&90k&5X&4.5X\\
inception\_4c &510k&102k&5X&4.3X\\
inception\_4d &605k&121k&5X&4.5X\\
inception\_4e &868k&174k&5X&4.3X\\
inception\_5a &1M&200k&5X&4.5X\\
inception\_5b &1M&200k&5X&4.5X\\\hline
fc8 &1M&200k&5X &5X \\\hline
Total &7M&1.3M&\textbf{5.4X} & 4.5X\\\hline
\end{tabular}

\label{table:compression_google_layer}
\end{table}

\begin{table}[t]
\centering
\vspace{-1em}
\caption{Comparisons of different approaches on GoogLeNet. $\#W$: The number of parameters. MCR: Maximum compression rate.}
 \begin{tabular}{c|c|c}\hline
Network	& $\#$W &MCR  \\ \hline
GoogLeNet &7M&1X\\
KD	& 5.1M&	1.4X\\
Low-Rank~\cite{tai2015convolutional}&2.4M& 2.8X\\
Tucker~\cite{kim2015compression}&4.7M& 1.3X \\
GreBdec~\cite{yucompressing} &1.5M&4.5X\\
Sparse~\cite{mao2017exploring}&2.3M &3X\\
DeepRebirth~\cite{li2017deeprebirth}&2.8M&2.5X\\
\pname{} & 1.3M &5.4X\\
 \pname{}+KD &0.95M &7.4X \\\hline
\end{tabular}
\label{table:comparison_googlenet}
\vspace{-1em}
\end{table}

\section{Experiment}
In this section, we present the comprehensive evaluation results and analysis of the proposed \pname{} deep model compression method. Following previous model compression work~\cite{kim2015compression,yucompressing}, our evaluation is performed on two popular image classification CNN models pre-trained on ImageNet~\cite{imagenet}, namely GoogLeNet~\cite{googleNet} and VGG-16~\cite{vgg}. 
We also provide results of comparison with other state-of-art model compression approaches. We implement our method and conduct the experiments using Tensorflow~\cite{tensorflow}. Our pre-trained reference models are obtained from the model repository of the official Tensorflow release.

\subsection{Experimental Settings}
In our experiment, we compress up to four neural network layers at a time. The $X$ and $Y$ of a neural network layer are collected by feeding a training example and extracting the layer's input feature maps (as $X$) and output feature maps (as $Y$). To optimize a decomposition that the output feature maps from the decomposed layers 
have minimal reconstruction error which is critical for initialization of fine-tuning, we use 3000 input-output pairs per layer (i.e., 12,000 examples for four layers.)
Different from pruning-based model compression methods~\cite{han2015learning,kim2015compression,tai2015convolutional} which heavily rely on layer-wise fine-tuning, our approach allows fine-tuning of the complete model which is much more time efficient and it converges in merely two epochs. We let $\lambda = 10^{\eta}E_{max}(\frac{1}{n}(XX^T))$ ($\eta \in (-3, 3)$, $E_{max}$ is the largest singular value)~\cite{yucompressing}, $\lambda_2 = 0.13$ and $\lambda_{\theta} = 10^{-3}$ in all experiments.

To compare with state-of-the-art model compression methods, we use the \textbf{\textit{maximum compression rate}} (\textbf{\textit{MCR}}) as our evaluation metric. Specifically, we would like to compress a model to the minimum number of parameters while preserving its accuracy (without accuracy loss compared with original model). For ImageNet, we use the Top-1 and Top-5 testing accuracy on ILSVRC2012 validation dataset as a guidance to measure the \textbf{\textit{MCR}}, i.e., we make sure there is no accuracy loss on both Top-1 and Top-5 validation accuracy after model compression. We report the number of parameters and MCR to evaluate the efficiency of different compression approaches.

\begin{table}[t]
\centering
\vspace{-1em}
\caption{Compression statistics of \pname{} on VGG-16. $\#W(O)$: total parameters of original model. $\#W(C)$: total parameters of compressed model. R: compression rate.}
 \begin{tabular}{c|c|c|c|c}
\hline
Layers & $\#$ W (O) & $\#$ W (C)& R &GreBdec\\ \hline
conv1\_1 &2k&1k&2X&1X\\
conv1\_2 &37k&7k&5X&5X\\
conv2\_1 &74k&15k&5X&4.3X\\
conv2\_2 &148k&30k&5X&4.3X\\
conv3\_1 &295k&59k&5X&4.2X\\
conv3\_2 &590k&118k&5X&4.5X\\
conv3\_3 &590k&118k&5X&4.5X\\
conv4\_1 &1M&200k&5X&4.2X\\
conv4\_2 &2M&400k&5X&4.5X\\
conv4\_3 &2M&400k&5X&4.5X\\
conv5\_1 &2M&400k&5X&4.5X\\
conv5\_2 &2M&400k&5X&4.5X\\
conv5\_3 &2M&400k&5X&4.5X\\\hline
fc6 &103M&4.8M&21.6X&25X\\
fc7 &17M&0.8M&21.3X&25X\\
fc8 &4M&1M&5X&5X\\\hline
Total&138M&9M&\textbf{15.4X}&14.2X\\\hline
\end{tabular}
\label{table:compression_vgg_layer}
\vspace{-1em}
\end{table}

\begin{table}[t]
\centering
\caption{Comparisons of different approaches on VGG-16. $\#W$: the number of parameters. MCR: Maximum compression rate.}
 \begin{tabular}{c|c|c}\hline
Network	& $\#$W & MCR \\ \hline
VGG-16 &138M&1X\\
KD&	107M & 1.3X	\\
Low-Rank~\cite{tai2015convolutional}&50.2M& 2.8X\\
Tucker~\cite{kim2015compression}&127M& 1.1X\\
Pruned~\cite{han2015learning}&10.3M& 13.4X \\
GreBdec~\cite{yucompressing}&9.7M &14.2X\\
Bayesian~\cite{louizos2017bayesian}& 9.9M &14X\\
Sparse~\cite{mao2017exploring}&37M&3.7X\\
\pname{}&9M&15.4X\\
\pname{} + KD & 7.3M&18.9X\\\hline
\end{tabular}
\vspace{-1em}
\label{table:comparison_vgg}

\end{table}


\textit{\textbf{Implementations: GoogLeNet}} is composed of convolutional layers and one final fully-connected layer. For convolutional layers: there are 4 different filter sizes: 1x1, 3x3, 5x5 and 7x7. We compress each filter size separately. Note that for two filters with the same receptive field (e.g., 3x3), their depths are likely to be different depending on the feature map channels generated by its previous layer. In \pname{}, we want to learn a common components matrix $\hat{V}$ across multiple layers, separating multiple layers by different filter sizes is also necessary, so that the dimension of the common component matrix $\hat{V}$ can be unified by the fixed filter size. Therefore, to compress weight matrices among different layers, we set the depth of the common components as the greatest common divisor (GCD) of depth of the filters with the same receptive field size (among 4 nearby filters in our experiment), and thus each filter may be represented by one or more components. For example, a 3x3x48 filter may be represented by three 3x3x16 components. The final fully-connected layer is processed separately.

\textit{\textbf{VGG-16}} is composed of 13 convolutional layers and 3 fully-connected layers. All convolutional layers have the same 3x3 filter size. We follow the same GCD strategy to set the size of the common components. The convolutional layer 1 is compressed separately since its filter depth is only 3. The final 3 fully-connected layers are individually compressed since those weight matrices have different dimensions.

\subsection{Experimental Results}
Table~\ref{table:compression_google_layer} and Table~\ref{table:compression_vgg_layer} give the details of compression rate within each layer for GoogLeNet and VGG-16, respectively. In addition, we set three baseline methods: (1) the original pre-trained model, (2) Knowledge Distillation (KD)~\cite{hinton2015distilling} which trains a student network model with the same network architecture of the final compressed model, and (3) GreBdec~\cite{yucompressing}, the single-layer model compression in Eq.~\ref{eq:single-layer}. Comparisons with other works on GoogLeNet and VGG-16 are listed in Table~\ref{table:comparison_googlenet} and Table~\ref{table:comparison_vgg} respectively. Obtained results demonstrate that the proposed approach has achieved largest compression rate without accuracy loss.

Table~\ref{table:compression_google_layer} and Table~\ref{table:comparison_googlenet} give detailed compression statistics and the comparisons on GoogLeNet. We can compress GoogLeNet by $5.4$ times with higher accuracy compared to $4.5$ times achieved by GreBdec~\cite{yucompressing}. Furthermore, compared to \cite{yucompressing}, we achieve a better compression rate on each convolutional layer (on average more than $20\%$) as indicated in Table~\ref{table:compression_google_layer}. The advantage is even more significant for the first few layers (conv1\_1, conv2), for which the number of parameters can be reduced by more than 50\% compared to \cite{yucompressing}. These results demonstrate that mining cross-layer inherent knowledge can improve the power of compression on deep neural networks. 
The compressed GoogLeNet model can be further compressed using knowledge distillation. Specifically, we apply KD by using the compressed model obtained from \pname{} as the teacher model and we gradually remove the inception layers from the teacher model as the student model. We still use the Top-1 and Top-5 accuracy to guide KD until accuracy loss is observed. In the experiment, we remove inception\_5a and inception\_5b to get the student model which can further reduce the number of weight parameters to less than 1M (7.4x MCR). This result demonstrates that by combining \pname{} with KD, the model can be further compressed to better satisfy the need of mobile deployment. 


Table~\ref{table:compression_vgg_layer} and Table~\ref{table:comparison_vgg} give detailed compression statistics and the result comparison on VGG-16. The MCR obtained on VGG using \pname{} is 15.4x compared to 14.2x achieved by the state-of-the-art method GreBdec~\cite{yucompressing}. Louizos \etal~\cite{louizos2017bayesian} used the variational Bayesian approximation compressed VGG-16 to 14X while we can achieve 15.4X. Compared to GoogLeNet, fully-connected layers constitute the majority redundancies of VGG-16. Therefore, since we can only compress the fully-connected layers individually, the proposed \pname{} method shows less advantage over existing methods compared to the result of GoogLeNet (Table~\ref{table:compression_google_layer} ). Furthermore, we can get 18.9x MCR by fusing with KD (student model learns smaller fc6 and fc7).

\begin{figure}
\centering
\vspace{-1em}
\includegraphics[height=4cm]{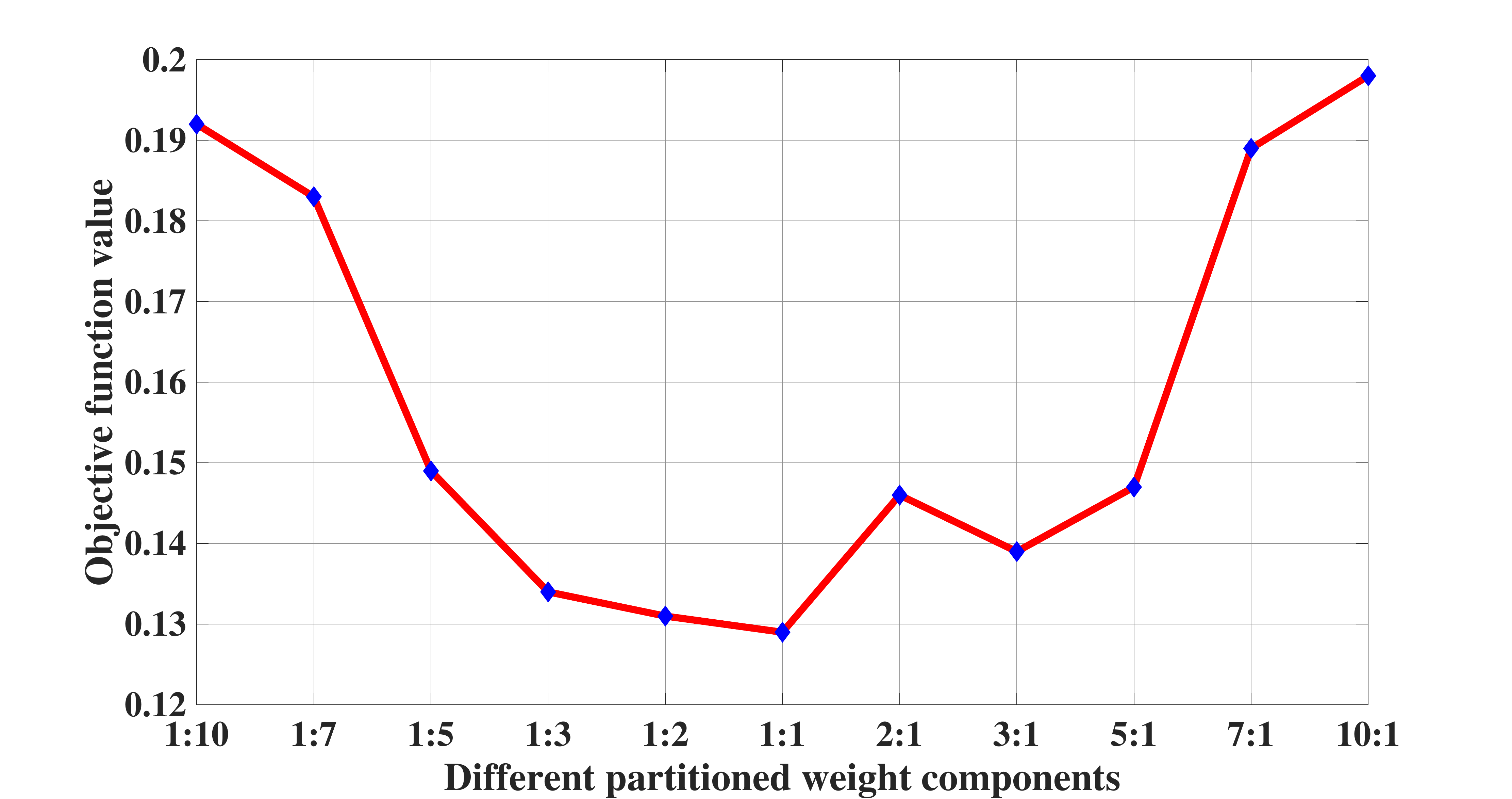}
\caption{Illustrations of performance influence caused by varying the percentage of common weight components $\hat{V}$ on VGG-16.}
\label{fig:3}
\vspace{-1em}
\end{figure}

\subsection{Impact Factor Analysis in \pname{}}

\begin{table}[t]
\centering
\caption{The results of considering inherent similarity knowledge. FCR: Fix Compression Rate, Top-1/Top-5 Accuracy Drop Percentage}
 \begin{tabular}{c|c|c|c}\hline
Network	& FCR & Top-1 $\downarrow$ & Top-5 $\downarrow$\\ \hline
MIC GoogLeNet &5.4X &0.21\%&0.33\%\\
\textbf{ \pname{}} GoogLeNet &\textbf{5.4X}& \textbf{0\%} & \textbf{0\%}\\\hline
MIC VGG-16& 15.4X& 0.37\%&0.23\%\\
\textbf{\pname{}} VGG-16&\textbf{15.4X}&\textbf{0\%}&\textbf{0\%}\\\hline
\end{tabular}
\label{table:MIC}
\vspace{-1em}
\end{table}

\textbf{Mining Inherent Similarity Knowledge.} We study the influence of enforcing the inherit similarity knowledge to the compression pipeline and present the result in Table~\ref{table:MIC}. In this experiment, we try to compress the models to the same number of parameters reached by the MCR of \pname{} but without adding the inherent similarity term $\theta$ (we name this method as MIC), which corresponds to the loss function defined in Eq.~\ref{eq:6}. As shown in Table~\ref{table:MIC}, both GoogLeNet and VGG-16 models have some accuracy loss. Therefore, adding inheriting similarity constraint can help maintain the model accuracy. Without adding this constraint, the training process cannot distinguish distant layers from nearby layers, which results in ambiguities. 


\textbf{Ratio of Common and Individual Components.} As discussed above, one major contribution of this work is the proposed component sharing across layers. It is necessary to evaluate the initialization of appropriate rate of weight sharing components among layers. We use the smallest weight matrix as our base matrix and use the smallest $m_t$ as the upper bound of the common components. Then, we set different ratios of common components with regard to all weight components. We set the initial ratio to 1:10 and slowly increase the number of shared common components to 1:7, 1:5, 1:3, 1:2, 1:1, 2:1, 3:1, 5:1, 7:1 and 10:1. We use convolutional layer 2 to layer 13 in VGG-16 as the input of \pname{} and the results are evaluated by the value of objective function in Eq.~(\ref{eq:5}). A smaller value indicates smaller difference between the original weight matrices and the compressed matrices for retraining, and thus gives a better initialization for retraining. In our experiments, similar to \cite{yucompressing}, we observe that a good initialization is essential for recovering the model accuracy. Fig.~\ref{fig:3} shows the results of different partitioned weight components. The best result is obtained after splitting the weight components equally between common and individual. Based on this observation, we use the 1:1 as the ratio between common and individual weight components in all experimental settings. 

\textbf{Inference Speed.} In this paper, we focus more on saving the storage of deep learning models which has significant impacts, e.g., models can be distributed and deployed on mobile devices efficiently by saving both network bandwidth and mobile storage. Though inference speed was not discussed, \pname{} has great potential for substantial speed-up. In Eq.~\ref{eq:single-layer}, $Y = (L + S)X = LX + SX$, i.e., the inference on the layer is broken down into two parallel computations $LX$ and $SX$, both of which can be efficiently computed: (1) the low-rank matrix $L$ is decomposed into two small matrices $U$ and $V$ and the inference time is greatly reduced by calculating $VX$ before multiplying $U$, (2) for the sparse matrix $S$, the higher sparsity rate enables more efficient usage of sparse matrix-vector multiplication operators~\cite{liu2013efficient} for speed-up and a LASSO regression based pruning~\cite{he2017channel} has been proved help accelerate the inference time; in addition, novel hardware such as specialized neural computing chips have been proposed to further speed up sparse matrix computation~\cite{han2016eie}. Further investigation of the inference speed will be our future work.


\vspace{-0.5em}
\section{Conclusions and Future Work}
In this work, we propose a new deep model compression algorithm termed MICIK, which compresses deep neural networks by integrating intra and inter layer inherent similarity knowledge. Extensive experiments on ImageNet demonstrate the evident superiority of MICIK over the state-of-the-arts. In the future, we will investigate adaptive adjustment of the shared components among different layers. 

{\small
\bibliographystyle{ieee}
\bibliography{cvpr18}
}

\end{document}